\documentclass[12pt]{spieman}  
\usepackage{amsmath,amsfonts,amssymb}
\usepackage{graphicx}
\usepackage{setspace}
\usepackage{tocloft}
\usepackage{booktabs}
\usepackage{subcaption}
\usepackage{comment} 
\usepackage[table]{xcolor}

\title{Confidence-Based Annotation Of Brain Tumours In Ultrasound}

\author[a]{Alistair Weld}
\author[b]{Luke Dixon}
\author[a]{Alfie Roddan}
\author[c]{Giulio Anichini}
\author[c]{Sophie Camp}
\author[a]{Stamatia Giannarou}
\affil[a]{Hamlyn Centre, Imperial College London, UK}
\affil[b]{Department of Imaging, Charing Cross Hospital, UK}
\affil[b]{Department of Neurosurgery, Charing Cross Hospital, UK}

\cftpagenumbersoff{figure}
\cftpagenumbersoff{table} 
\begin{document} 
\maketitle

\begin{abstract}
\textbf{Purpose:} An investigation of the challenge of annotating discrete segmentations of brain tumours in ultrasound, with a focus on the issue of aleatoric uncertainty along the tumour margin, particularly for diffuse tumours. A segmentation protocol and method is proposed that incorporates this margin-related uncertainty while minimising the interobserver variance through reduced subjectivity, thereby diminishing annotator epistemic uncertainty.
\textbf{Approach:} A sparse confidence method for ultrasound data annotation is proposed, based on a protocol designed using computer vision and radiology theory.
\textbf{Results:} Output annotations using the proposed method are compared with the corresponding discrete annotation variance between the annotators. A linear relationship was measured within the tumour margin region, with a Pearson correlation of 0.8. The downstream application of image segmentation was explored, comparing training using confidence annotations as soft labels with using the best discrete annotations as hard labels. In all evaluation folds, the Brier score was superior for the soft-label trained network.
\textbf{Conclusion:} A formal framework was constructed to demonstrate the infeasibility of discrete annotation of brain tumours in B-mode ultrasound images. Subsequently, a method for sparse confidence-based annotation is proposed and evaluated. 
\end{abstract}


{\noindent Corresponding Author: \linkable{a.weld20@imperial.ac.uk} }

\begin{spacing}{2}   

\section{Introduction}
\label{intro}


The annotation of objects of interest in medical images is invaluable from both a clinical and technical perspective. For normal anatomy or pathological tissue, the annotation of these areas can serve a multitude of purposes that can include helping to make treatment decisions (for example, long-term monitoring) \cite{Folio2015QuantitativeRR}, or postoperative evaluation (for example, the extent of resection)\cite{Helland2023SegmentationOG}. In clinical practice, tumours can be annotated in many ways, such as diameter measurement \cite{Lura2022WhatMT} or arrows for localisation \cite{Steiger2016ProstateMB}. The most precise form of annotation for differentiating pathological tissue from normal tissue is discrete segmentation \cite{Menze2015TheMB}. This involves the binary separation of the tumour from its surroundings/background along the boundary of the tumorous area or separation via different categories of the tumour and surrounding normal tissue, such as enhancing, non-enhancing, oedema and necrosis.

In the clinical setting, ultrasound (US) is one of the most widely used imaging modalities. The NHS publication ``Diagnostic Imaging Dataset Annual Statistical Release 2022/23'' \cite{Dixon_2023} reports that between the years 2012/13 to 2022/23, US imaging accounted for $>20\%$ of all recorded imaging activity, the second most commonly used modality behind X-rays, where the common application of US was for the abdomen/pelvis. 

\begin{figure}[t]
\centering
\includegraphics[width=\columnwidth]{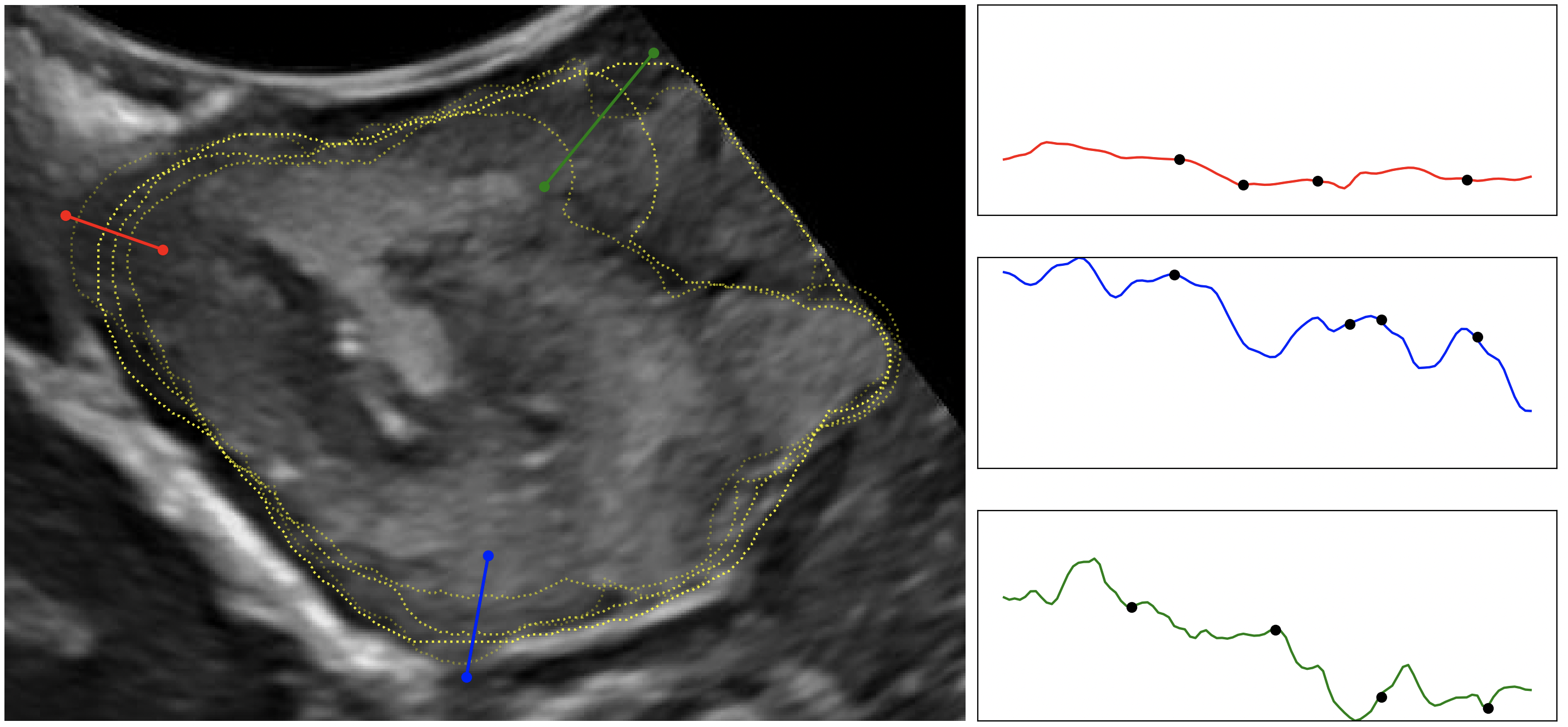}
\caption{Tumour margin delineation variability among four professional annotators. The three vectors (red, green, blue) show visible ambiguity over the decision of the edge of the tumour, where the segments are shown with yellow dotted lines of varying transparency.}
\end{figure}

The importance of US has resulted in published single- and multicentre data, which has been beneficial for both clinical education and technical research. In particular, motivated by recent successes with machine learning, annotated datasets have been curated for the task of semantic segmentation. Some examples of such are breast cancer \cite{ALDHABYANI2020104863}, cardiac structures \cite{8649738}, and foetus head \cite{fhead}. A common denominator of these datasets is the clarity of the object of interest, where the boundaries of the objects are well-defined with limited visual uncertainty.

Intraoperative ultrasound (IOUS) has become a commonly used tool to assist in surgical resection, by localising pathological tissue and delineating its margins \cite{Lubner2021DiagnosticAP}. Recent evidence suggests that IOUS might be particularly useful in improving the likelihood of maximal safe resection during brain tumour surgery \cite{Dixon2022IntraoperativeUI}. Although a useful tool, due to the limited standardisation of specific clinical training and the difficulty of interpreting images due to the confined field of view, the annotation of brain tumours in IOUS images can be difficult, resulting in variance between observers \cite{Weld2023ChallengesWS}. This is further compounded in the case of diffuse tumours, where there is often an extremely poor separability of the tumour from the normal surrounding tissue due to the graduated infiltrative nature of the disease and the inherent lack of a boundary \cite{wen2017response, weller2021eanom, nagashima2006dural, berghoff2013invasion, beutler2024intracranial}. 

This article explores probabilistic annotation of brain tumours as an alternative to discrete annotation. Diffusion, and infiltration of a tumour and ill-definition of its margin ultimately make discrete annotation impossible. What is argued in this article is that annotation should be performed not on a pixel level but rather spatially and locally, in the form of distributed continuous confidence values. The key contributions of this work are as follows:
\begin{itemize}
    \item We propose a confidence-based annotation protocol and method.
    \item We verify that training tumour segmentation networks with clinically relevant soft labels improves neural network calibration compared to training with hard labels.
\end{itemize}

\section{Methodology}
\label{MaM}

\subsection{Annotation Protocol}

Confidence maps are the proposed alternative annotation method, compared to standard segmentation. Using heat map distributions to represent the perceptual likelihood/confidence in the tumour. This section describes a guideline for how annotators should interpret images and perform the annotation. Fundamental inspiration has been taken from other radiological annotation approaches that differentiate the different layers of the tumour and the relevant surrounding tissue \cite{Menze2015TheMB}. Specifically, for the construction of the confidence map of the object, the annotation utilises three grades of scatter points - high, medium, and low confidence. Each scatter point is used to delimit the edge of the corresponding level of confidence over which a heat map is to be constructed/fitted. This protocol encourages sensitivity over specificity of the annotation  to account for the the impact of aleatoric uncertainty along the tumour margins. Fig.~\ref{fig:gtgt} displays examples of ground truth annotation using the graded scatter points.

Areas of high confidence can be perceived as where the tumour mass is well defined with expected low levels of interobserver annotation variance - low aleatoric uncertainty in the pixel-level. In these areas, the granularity of US is sufficient to make the decision on whether to resect or not. This can be contributed to the area belonging to the hypercellular components of the tumour, which is usually hyperechogenic and thus considerably brighter than it's surroundings. Furthermore, where the texture is anomalous, significantly homogeneous or heterogeneous relative to the surrounding normal tissue. Or, where edges are well defined by calcification or necrosis.

The medium confidence annotation should primarily be associated with the edge of the fuzzy margin. Here there is likely a large degree of interobserver variability, and the aleatoric uncertainty is already high. For these zones, the decision to resect is difficult to make using purely the US image as a reference, where granularity begins to become or is unreliable. In particular, for a diffuse tumour, this is somewhere along the gradient between a clear tumour core and perceived normal tissue. For a well-defined tumour, this can be areas around the clear margin of the tumour that have the continued visual properties of the tumour. For example, continued brightness and or continued anomalous texture. This can also be due to obscuration or acoustic shadowing. This level of confidence is likely the hardest to annotate.

The distribution of low-confidence annotations will occur sparsely outside of the clear tumour mass and the ``clear fuzzy margin''. The aleatoric uncertainty and interobserver annotation variance are significant in these areas. The US granularity is not sufficient to make actionable decisions on resection. These areas will likely reflect the surrounding abnormal tissue that is a variable combination of oedema and less cellular tumour infiltration that correlates with the perilesional, non-enhancing abnormal signal observed on MRI \cite{Wrtemberger2022DifferentiationOP}. Pragmatically, these would be the areas that would not typically be resected, as it is thought to reflect predominantly oedema, but where there remains an uncertainty and risk of residual tumour cells.

\subsection{Confidence Map Construction}

Once the annotation is complete, the following sets of annotation points are defined: high likelihood $\mathcal{C_H} = \{(x_a, y_a)\}_{a=1}^{n_H}$, medium likelihood $\mathcal{C_M} = \{(x_b, y_b)\}_{b=1}^{n_M}$, and low likelihood $\mathcal{C_L} = \{(x_c, y_c)\}_{c=1}^{n_L}$.

The mesh grid $\mathcal{G} = \left\{(x, y) \mid x \in \left[0, \frac{W}{\lambda}, \frac{2W}{\lambda}, \dots, W\right], \, y \in \left[0, \frac{H}{\lambda}, \frac{2H}{\lambda}, \dots, H\right] \right\}$ is then defined, given the width $W$ and height $H$ of the image, subsampled using $\lambda$. For this work $\lambda=100$ provided a sufficiently smooth map.

The mesh grid is then weighted by aligning annotation points with grid points.
\begin{equation}
w(x, y) = 
\begin{cases} 
\beta_H & \text{if } (x, y) \in \mathcal{DT}(\mathcal{C_H}), \\
\beta_M & \text{else if } (x, y) \in \mathcal{DT}(\mathcal{C_M}), \\
\beta_L & \text{else if } (x, y) \in \mathcal{DT}(\mathcal{C_L}), \\
0 & \text{otherwise.}
\end{cases}
\end{equation}
where, Delaunay triangulation $\mathcal{DT}$ is used to determine which grid points correspond to each likelihood region. $\beta_H, \beta_M, \beta_L$ were assigned values $1, 0.5, 0.1$, respectively. These parameters are tunable, the chosen values reflecting the likelihood weighting boundary range (0–1).

The confidence map $\mathcal{CM}$ is finally constructed by applying Gaussian Kernel Density Estimation (KDE) (defined by $f$), followed by min-max normalisation, over the mesh grid $\mathcal{G}$. A minimal bandwidth factor $\sigma=0.1$ was chosen to prevent the over-influence of high-confidence areas and further reduction of spatial specificity.
\begin{equation}
f(x, y) = \frac{1}{n\sigma^22\pi} \sum_{i=1}^n w(x_i, y_i) e^{-\frac{1}{2}(\frac{(x - x_i)^2 + (y - y_i)^2}{\sigma^2})}
\end{equation}

\begin{equation}
\mathcal{CM} =\frac{f(x, y) - \min(f)}{\max(f) - \min(f)},
\end{equation}
where, $n$ is the number of points in $\mathcal{G}$. Overall, the set of tunable parameters is $[\lambda, \beta_H, \beta_M, \beta_L, \sigma]$, which corresponds to [meshgrid size, annotation class weighting, KDE bandwidth].

\section{Experiments And Results}
\label{Results}

\begin{figure}[t]
    \centering
    \begin{subfigure}[t]{0.7\textwidth}
        \centering
        \includegraphics[width=\linewidth]{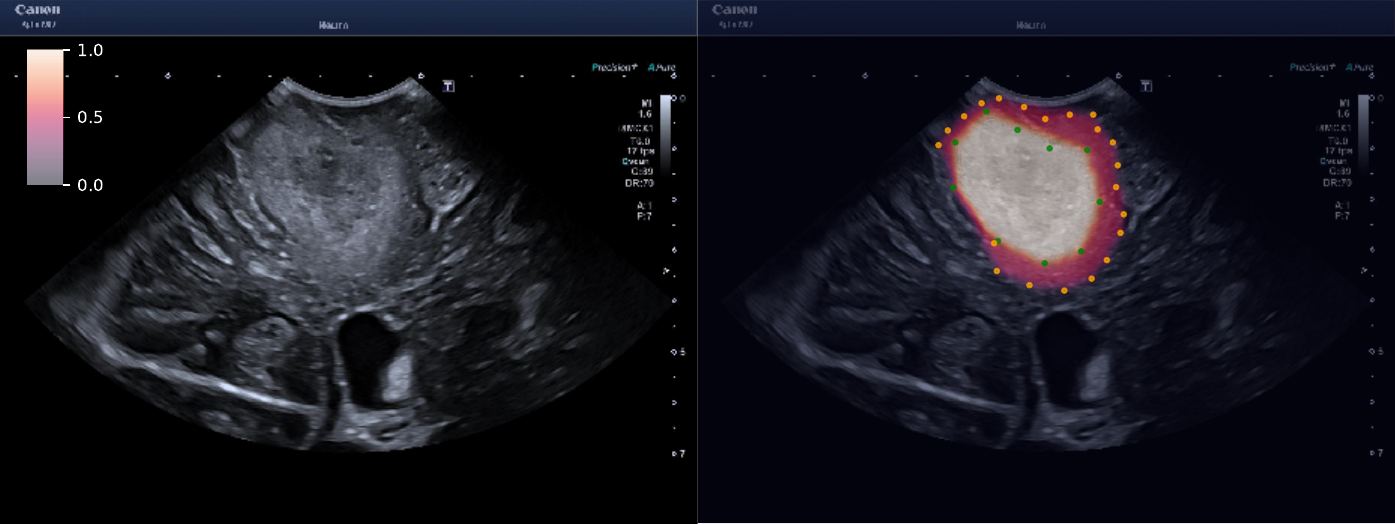}
    \end{subfigure}
    \begin{subfigure}[t]{0.7\textwidth}
        \centering
        \includegraphics[width=\linewidth]{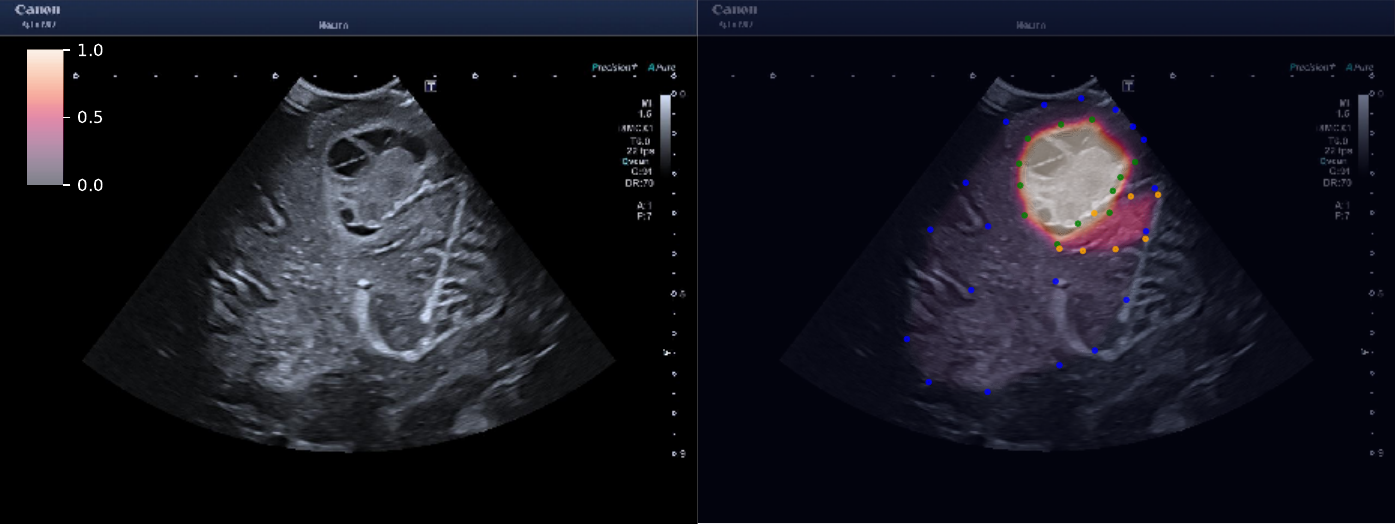}
    \end{subfigure}
    \begin{subfigure}[t]{0.7\textwidth}
        \centering
        \includegraphics[width=\linewidth]{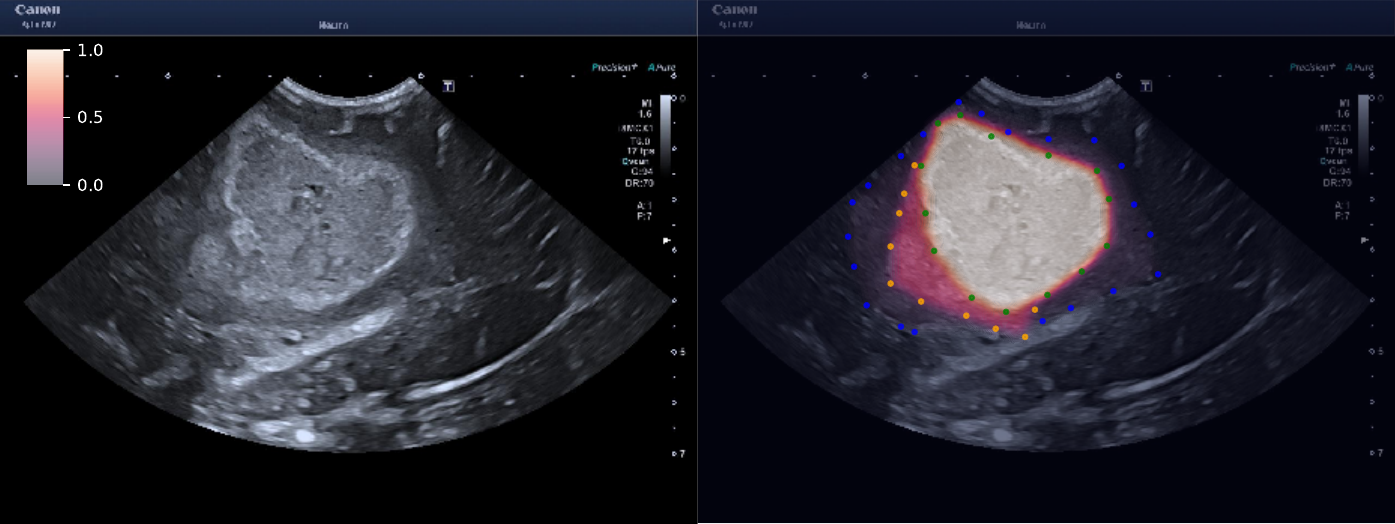}
    \end{subfigure}
    \caption{Sample ground truth annotations. The green points are points of high confidence, yellow are medium and blue are low.}
    \label{fig:gtgt}
\end{figure}

This section will describe the evaluation setup that has been used to validate the usability of this method. Through, the implicit prediction of interobserver variance as well as down stream data learning application. Sample annotations are displayed in Figure.~\ref{fig:gtgt}.

\subsection{Confidence Map And Interobserver Variance Similarity}

Data from \cite{Weld2023ChallengesWS} has been used in this study to measure the correlation of confidence annotation with interobserver variance. The confidence annotations are performed with reference to the neuroradiologists annotations, which is defined in the cited study as their ground truth due to annotation in conjunction with multiple modalities, with the annotations by the neurosurgeons unseen. The neuroradiologist annotations are used primarily for localisation, with the annotations strictly adhering to the defined protocols. In particular, this means that not all neuroradiologist annotations are described as high confidence. 

\subsubsection{Metric Similarity}\label{sec:gt_similarity}

To start, a basic image-wise comparison is performed between the confidence maps and the interobserver variance, denoted \textit{Agreement} = $\frac{A_{1} + A_{2} + A_{3} + A_{4}}{4}$, where $A$ is an individual binary annotation. Two metrics are used, the Brier score and the Soft Dice score. Using both metrics accounts for pixel-wise similarity and segment similarity.
\begin{equation}
    \text{Brier} = \frac{1}{n} \sum_{i=1}^{n} (p_i - y_i)^2
    \label{eq:brier}
\end{equation}

\begin{equation}
    \text{Soft Dice} = 1 - \frac{2 \sum_{i=1}^{n} p_i y_i}{\sum_{i=1}^{n} p_i^2 + \sum_{i=1}^{n} y_i^2}
    \label{eq:sd}
\end{equation}
where, n is the dataset image index. The Brier score is calculated for both the entire image (considering foreground and background) and exclusively for the foreground, where the foreground is defined as the union of the agreement and confidence map. The Brier score with background is 0.00762 and without 0.0437. The Soft Dice is calculated on the entire image, and an average score of 0.076 is recorded. 
Both metrics indicate a strong agreement between the confidence map and \textit{Agreement} maps.

\subsubsection{Similarity Across Threshold Levels}

\begin{figure}[th]
    \centering
    \includegraphics[width=0.495\textwidth]{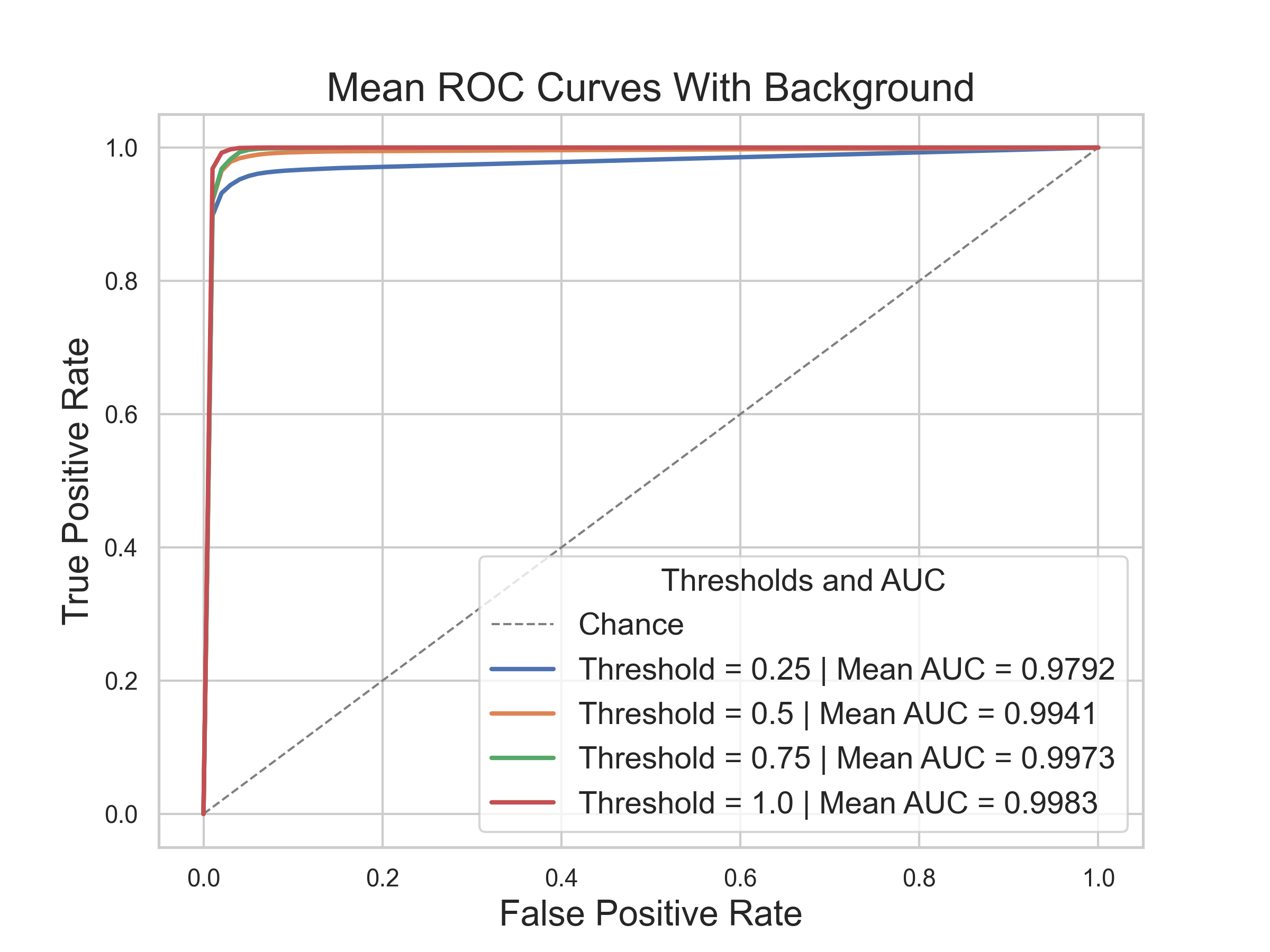}
    \includegraphics[width=0.495\textwidth]{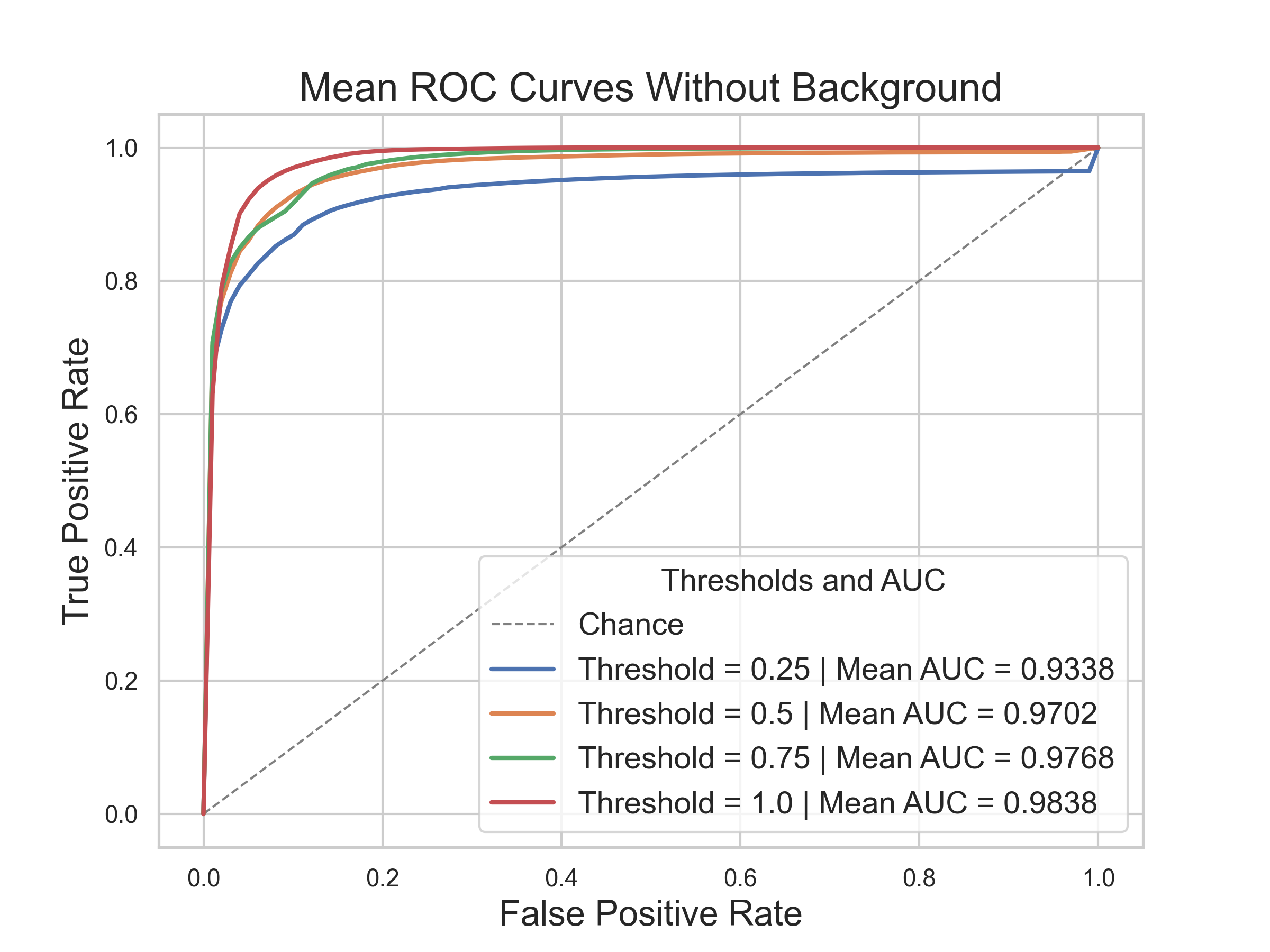}
    \caption{Mean Receiver Operator Characteristic (ROC) curves across threshold interobserver agreement levels. Left subfigure shows the ROC with all pixels included, while the right subfigure shows the ROC only using union area between the two annotations.}
    \label{fig:roc}
\end{figure}

To further assess how well the confidence maps align with the \textit{Agreement} maps, the Receiver Operating Characteristic (ROC) curve is computed at four different binary thresholds of the ground truth, $[0.25, 0.50, 0.75, 1.00]$. The ROC for across the dataset is calculated for both the entire image and exclusively for the foreground. The results of this ROC analysis are shown in Figure.~\ref{fig:roc}. For both cases, the Area Under the Curve (AUC) is greater than $0.9$, with the AUC improving as the threshold increases, which can be attributed to the reduced margin uncertainty. This result adds further evidence that the proposed protocol for creating confidence maps is effective in estimating interobserver variance.

\begin{figure}[th]
\centering
\includegraphics[width=0.75\columnwidth]{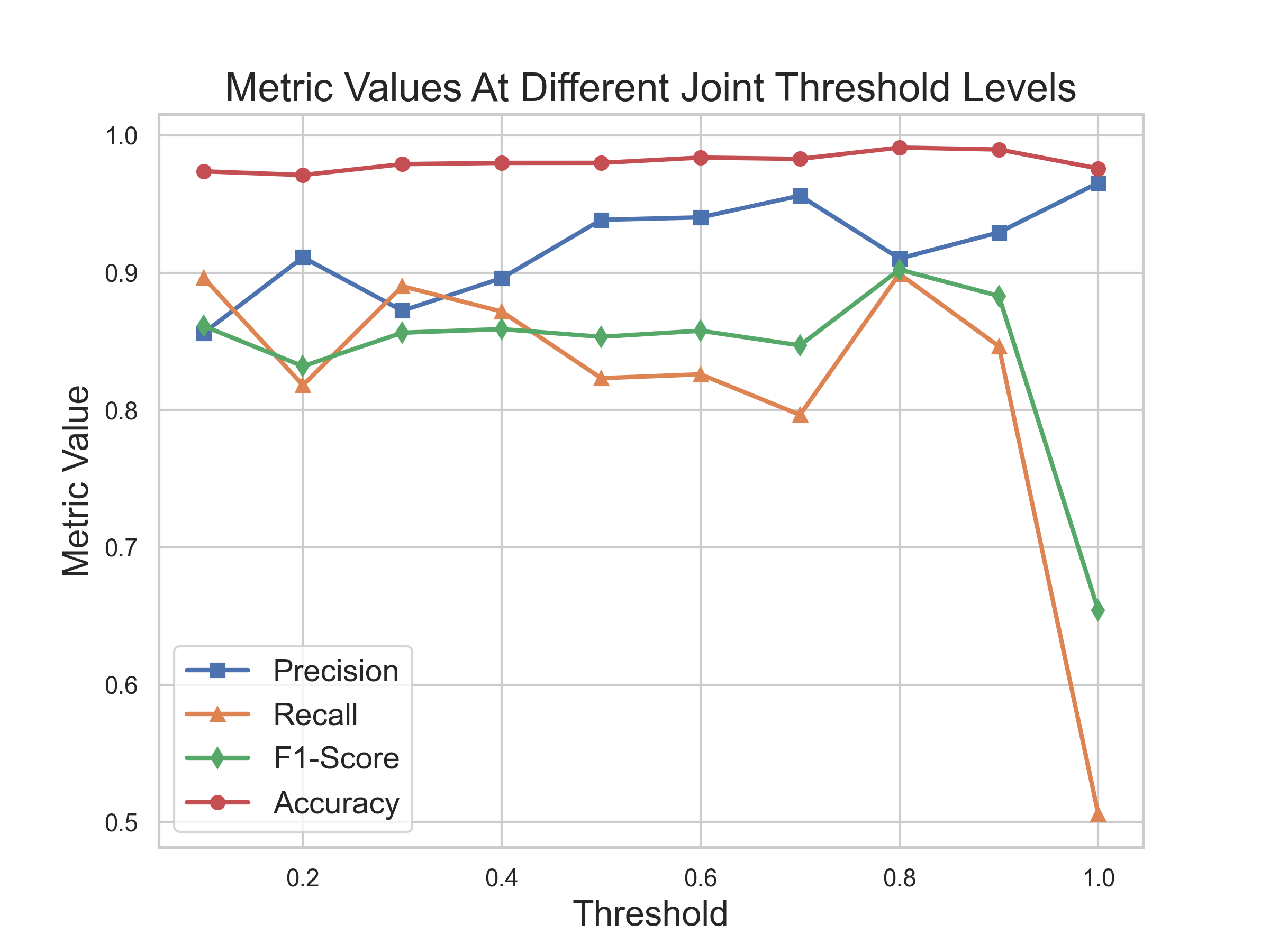}
\caption{Metric performance across different threshold levels, where the thresholds are applied to both annotations.}
\label{fig:doublethresh}
\end{figure}

Further threshold-based analysis is conducted, where in this instance, both the \textit{Agreement} and corresponding confidence map are jointly threshold binarised. This evaluation differs from the previous one by evaluating the similarity between the binarised confidence map and the binarised \textit{Agreement} map across multiple confidence levels. This joint thresholding approach effectively measures how well the confidence maps (and the annotation protocol) are calibrated to reflect the variability in interobserver agreement rather than just evaluating the overlap between annotations. The four evaluation metrics for this are Precision, Recall, F1-Score and Accuracy.




The results of this experiment are shown in Figure.~\ref{fig:doublethresh}, where is is notable that all metrics highlight good similarity except at the $1.0$ threshold, where there is reduced performance in Recall and F1-score. This can be attributed to an increased number of false negatives, due to the KDE's smoothness and therefore a reduced number of ones.

\subsection{Margin Uncertainty Correlation}

\begin{figure}[t]
\centering
\includegraphics[width=0.75\columnwidth]{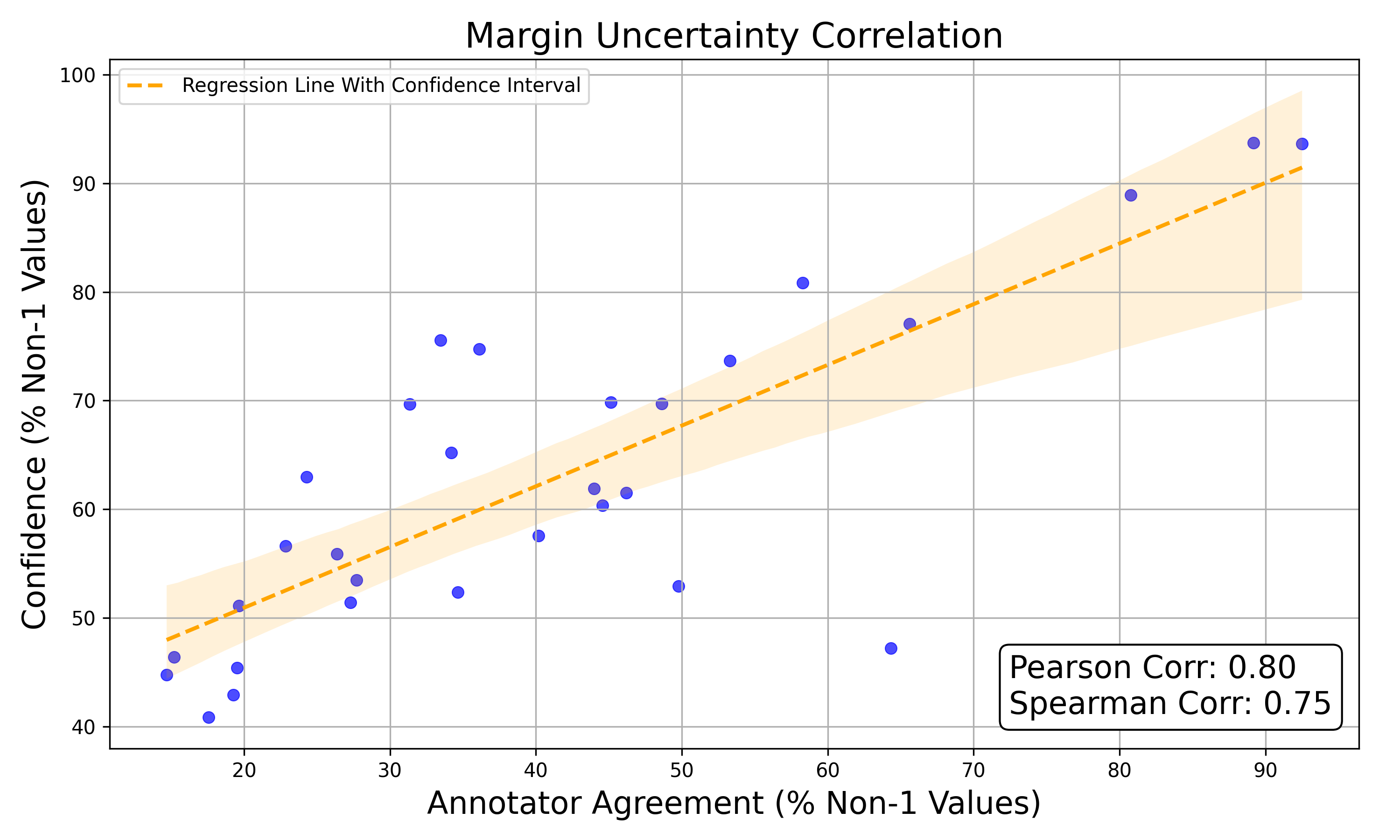}
\caption{The relationship between annotator variance and confidence annotation at the tumour margin (non-1 values) shows a clear linear trend, supported by correlating statistical evidence.}
\label{fig:pearspear}
\end{figure}

Extending the previous evaluation based on binarisation and segment analysis, analysis of the percentage of non-1 values in both confidence and interoberserver variance is conducted, focused on the margin uncertainty correlation between confidence map and \textit{Agreement} map. 
\begin{align*}
\psi_{\text{conf}} &= \frac{\left| \{ (x, y) \in I \mid 0 < \text{conf}(x, y) < 1 \} \right|}{\left| \{ (x, y) \in I \mid 0 < \text{conf}(x, y) \} \right|} \times 100 \\
\psi_{\text{agree}} &= \frac{\left| \{ (x, y) \in I \mid 0 < A(x, y) < 1 \} \right|}{\left| \{ (x, y) \in I \mid 0 < A(x, y) \} \right|} \times 100
\end{align*}
The metrics for this analysis are the Pearson correlation coefficient ($r$) and Spearman's rank correlation coefficient ($\rho_s$).
\begin{equation}
r = \frac{\sum_{i=1}^{n} \left( \psi_{\text{agree},i} - \overline{\psi_{\text{agree}}} \right) \left( \psi_{\text{conf},i} - \overline{\psi_{\text{conf}}} \right)}{\sqrt{\sum_{i=1}^{n} \left( \psi_{\text{agree},i} - \overline{\psi_{\text{agree}}} \right)^2} \sqrt{\sum_{i=1}^{n} \left( \psi_{\text{conf},i} - \overline{\psi_{\text{conf}}} \right)^2}}
\end{equation}

\begin{equation}
\rho_s = 1 - \frac{6 \sum_{i=1}^{n} d_i^2}{n(n^2 - 1)}
\end{equation}

The results are displayed in Figure.~\ref{fig:pearspear}, where the measured Pearson correlation was $0.80$ and the Spearman correlation $0.75$. The scatter plot shows the correlation between confidence map and interobserver variance, with a linear regression model fitted to the data with a 95\% confidence interval. The conclusion of this experiment is that the proposed annotation method will correctly increase the spread of confidence as the size of the tumour and its margins (and uncertainty) increase spatially.

What can be concluded from this set of evaluations is that there is a correlation between confidence maps and unseen interobserver variance. Not only does this affirm the benefit of this protocol and method of annotation in minimising the annotator-level epistemic uncertainty through treating the annotation as a local density spatial task rather than at a discrete pixel/voxel level. But also, these results suggest that this annotation strategy would greatly reduce time taken, as there is a reduced or even no requirement for multiple annotators.

\subsection{Downstream Application: Soft Label Learning}

The proposed annotation method is faster and simpler than traditional segmentation and can facilitate dataset generation. Confidence annotations, which cover a larger pixel area and incorporate likelihood grading, may provide more nuanced information about tumours. In deep learning segmentation tasks, probabilistic pixel-wise classification is typically performed using sigmoid or softmax activation functions, with discrete annotations. Studies, such as \cite{Szegedy2015RethinkingTI, Mller2019WhenDL, Vyas2020LearningSL} have explored the benefit of using soft labels, continuous labels in the binary case, and non-one-hot vectors in the multiclass case, for regularisation. The work proposed in \cite{Gros2020SoftSegAO} found that soft labels improved the sensitivity of segmentation networks, particularly for small objects in medical images.

\begin{figure}[htbp]
    \centering
    
    \begin{subfigure}[t]{0.49\textwidth}
        \centering
        \includegraphics[width=\linewidth]{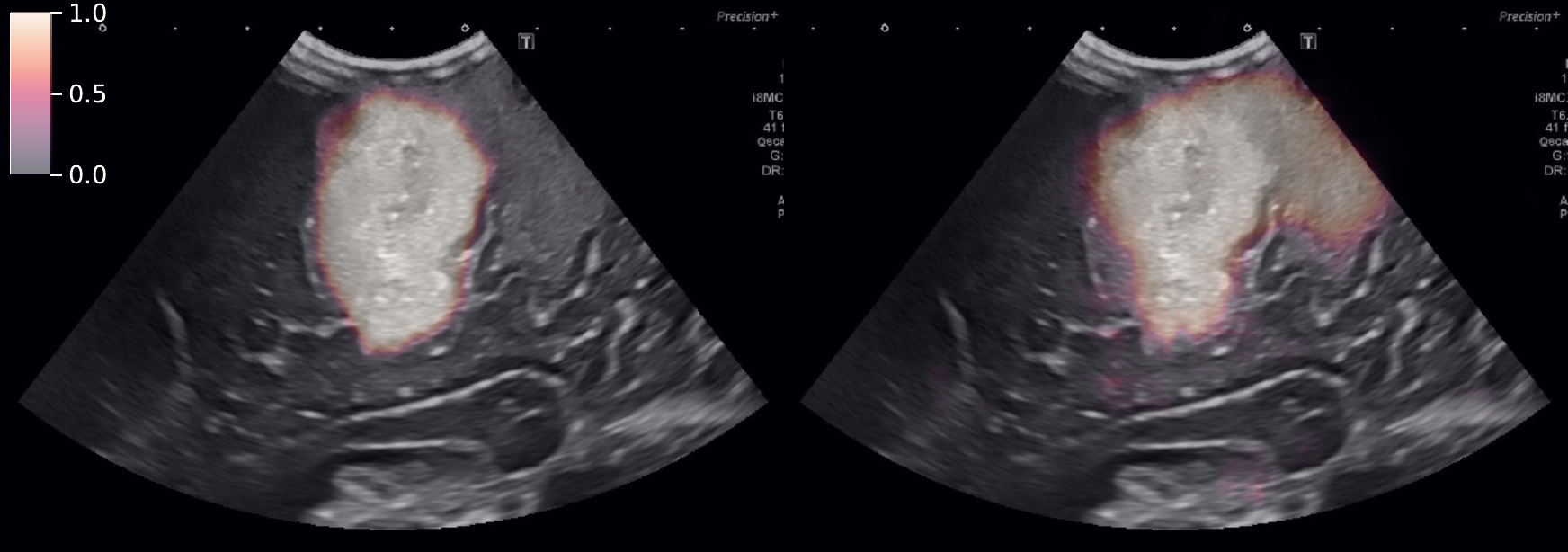}
    \end{subfigure}
    \begin{subfigure}[t]{0.49\textwidth}
        \centering
        \includegraphics[width=\linewidth]{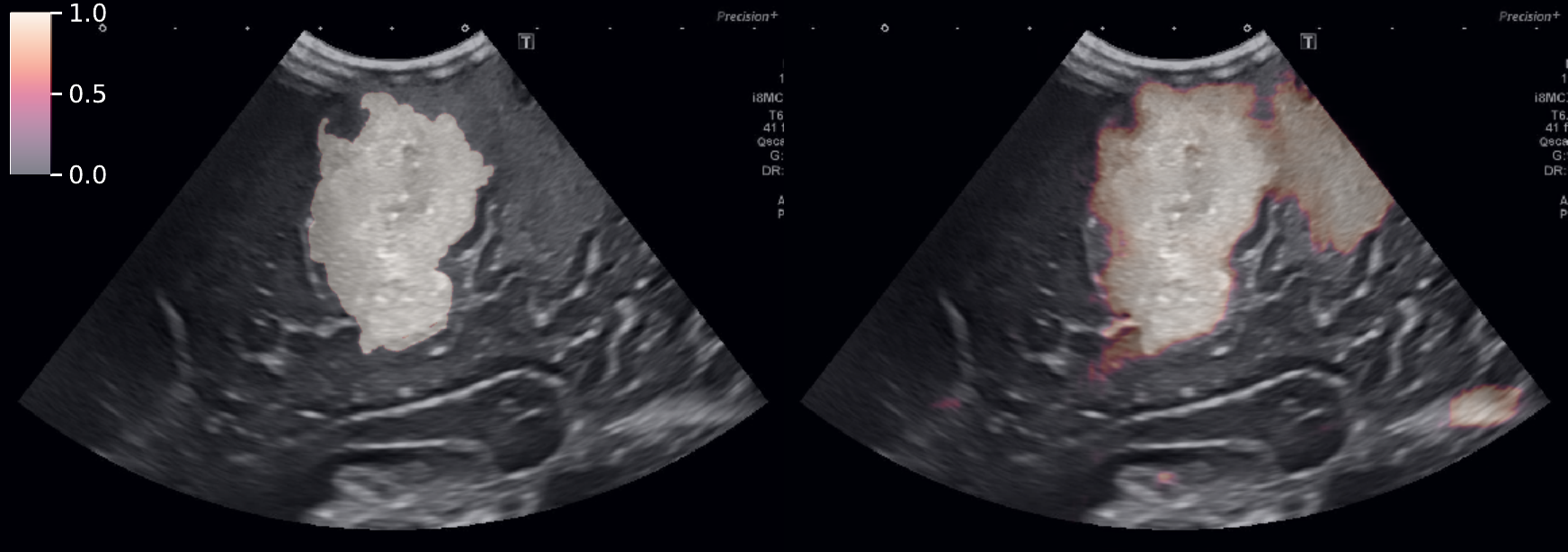}
    \end{subfigure}

    \begin{subfigure}[t]{0.49\textwidth}
        \centering
        \includegraphics[width=\linewidth]{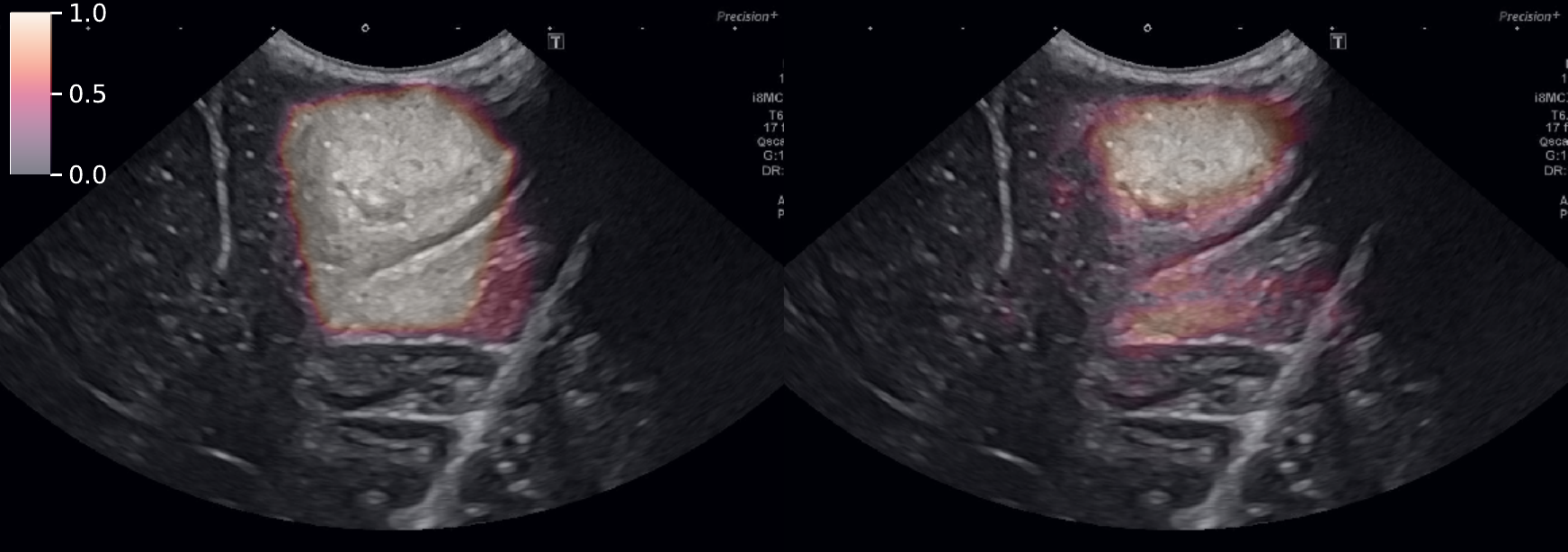}
    \end{subfigure}
    \begin{subfigure}[t]{0.49\textwidth}
        \centering
        \includegraphics[width=\linewidth]{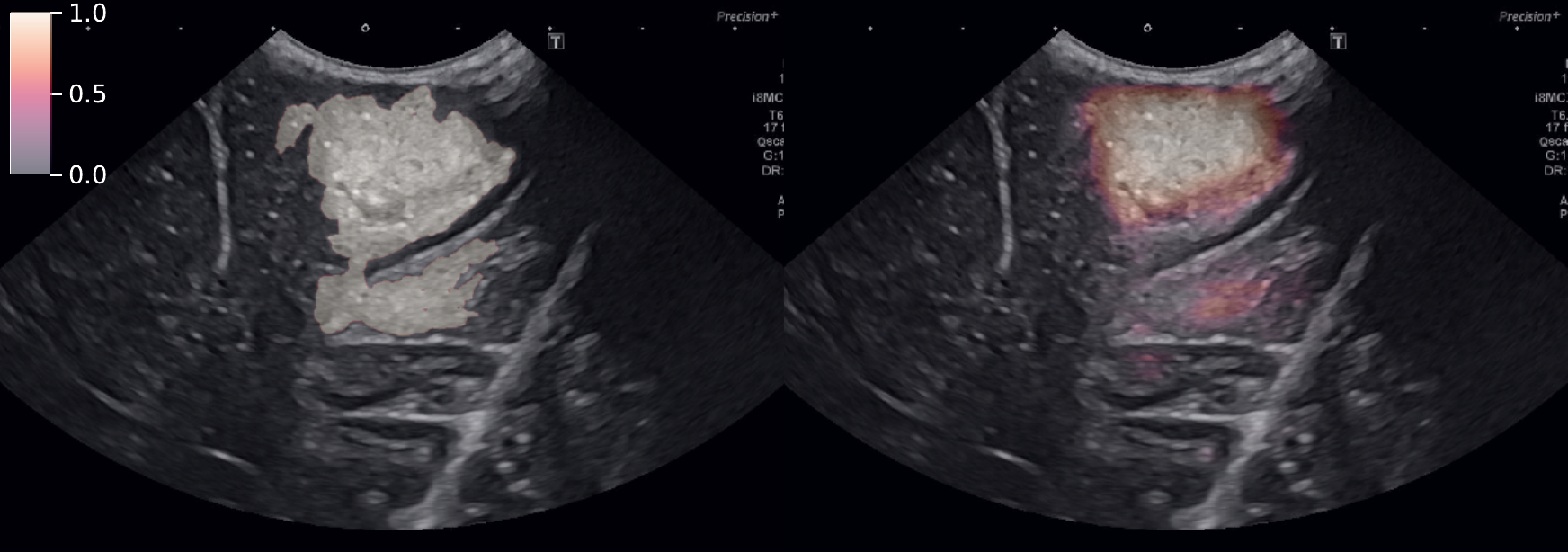}
    \end{subfigure}

    \begin{subfigure}[t]{0.49\textwidth}
        \centering
        \includegraphics[width=\linewidth]{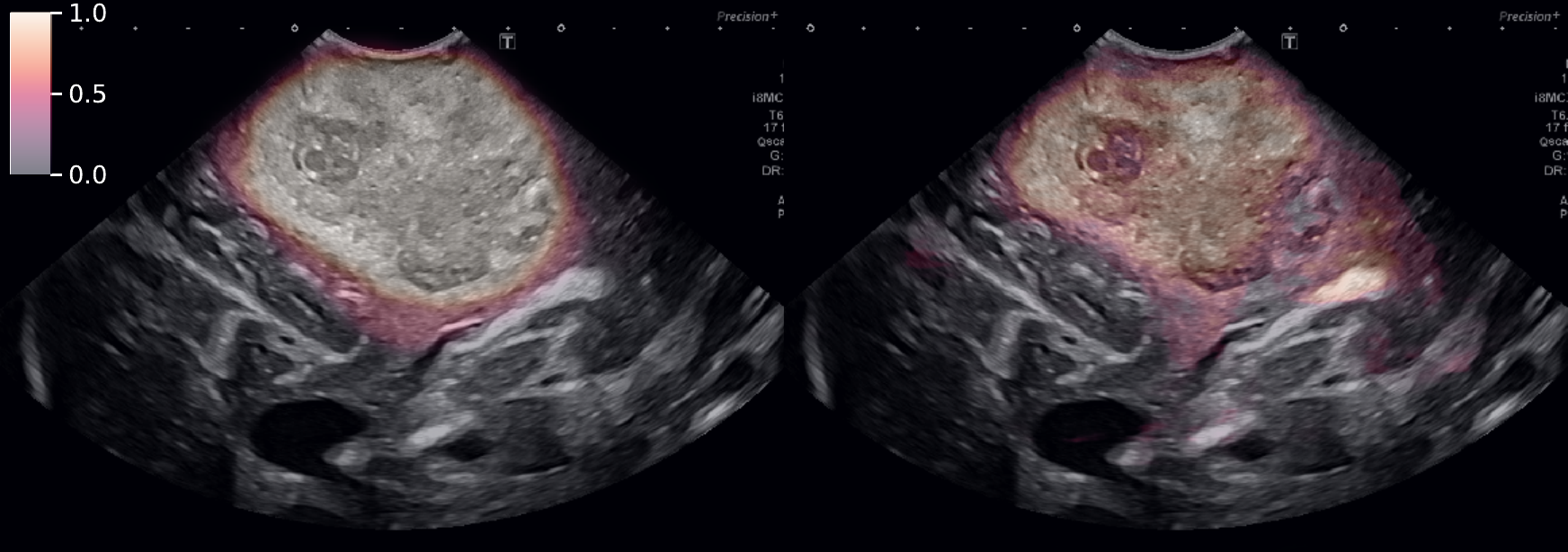}
    \end{subfigure}
    \begin{subfigure}[t]{0.49\textwidth}
        \centering
        \includegraphics[width=\linewidth]{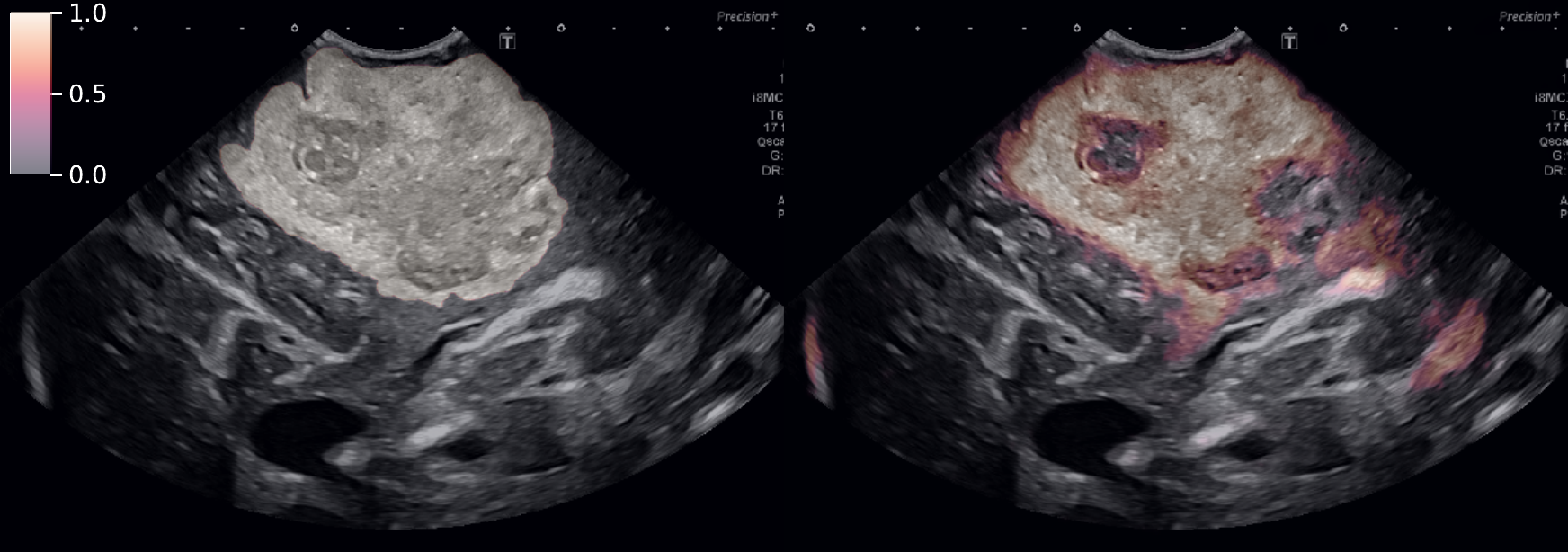}
    \end{subfigure}
    
    \caption{Test set output comparison from training on soft (left column) and hard (right column) ground truth. Where in each column, the left image is the ground truth and right the UNet output. }
    \label{fig:unet_op}
\end{figure}

In this work, experiments are performed by training a standard UNet model on both the confidence (soft) and discrete (hard) annotations, where the latter was the annotation from the neuroradiologist. This was done to determine the downstream benefit of this method for annotation against binary segmentation for the application of standard neural network training. For this experiment, a UNet model with 4 encoder/decoder layers and a maximum channel depth of 64 was trained using sigmoid activation in the final layer. Training was performed separately with the Brier and Soft Dice (Equation.~\ref{eq:brier},~\ref{eq:sd}) loss functions, as well as the following other loss functions, under a 5-fold cross-validation framework:
\begin{equation}
    \text{Binary Cross Entropy} = -\frac{1}{n} \sum_{i=1}^{n} \left[ y_i \log(p_i) + (1 - y_i) \log(1 - p_i) \right]
\end{equation}

\begin{equation}
    \text{Huber} = \frac{1}{n} \sum_{i=1}^{n} \begin{cases} 
      0.5 \cdot (p_i - y_i)^2, & \text{if } |p_i - y_i| < 1 \\
      |p_i - y_i| - 0.5, & \text{if } |p_i - y_i| \geq 1
   \end{cases}
\end{equation}

\begin{equation}
    \text{Tversky} = 1 - \frac{\sum_{i=1}^{n} p_i y_i}{\sum_{i=1}^{n} p_i y_i + \alpha \sum_{i=1}^{n} p_i (1 - y_i) + \beta \sum_{i=1}^{n} y_i (1 - p_i)}
\end{equation}
where, the \text{Tversky} parameters are $\alpha=0.3, \beta=0.7$ \cite{Salehi2017TverskyLF}. The metrics used for comparison are the same as in Section.~\ref{sec:gt_similarity}, where the $\text{Brier}$ score is particularly informative of calibration performance \cite{Brier1950VERIFICATIONOF}. Results of this are shown in Table.~\ref{tab:brier_scores}. Displayed in Figure.~\ref{fig:unet_op} is a sample of outputs for the $\text{Binary Cross Entropy}$ test set. Qualitatively, the soft trained network produces smoother confidence. Notably, more likely to activate with low confidence in high uncertainty regions. Unlike the hard trained network which tends to behave more discretely. Notably, the necrotic sphere in the bottom row is ignored by the hard trained network but low confidence is activated with the soft trained network. Which is a more desirable when considering that the necrotic sphere, in both networks, is surrounded by activated pixels.

\begin{table}[htbp]
    \centering
    \caption{Training a UNet on soft and hard annotations with varying loss functions. The results correspond to the average across the 5-folds. For all loss functions, the soft trained networks achieves better test performance than the hard trained networks.}
    \begin{tabular}{lccccc}
        \toprule
            Loss Function & \multicolumn{2}{c}{Brier} & \multicolumn{2}{c}{Soft Dice} \\
            \cmidrule(lr){2-3} \cmidrule(lr){4-5}
            & Soft & Hard & Soft & Hard \\
        \midrule
        Brier & \cellcolor{blue!25}0.0179 & 0.0234 & \cellcolor{blue!25}0.2219 & 0.2989 \\
        Soft Dice & \cellcolor{blue!25}0.0247 & 0.0270 & \cellcolor{blue!25}0.2236 & 0.2868 \\
        Binary Cross Entropy & \cellcolor{blue!25}0.0176 & 0.0220 & \cellcolor{blue!25}0.2132 & 0.2833 \\
        Huber & \cellcolor{blue!25}0.0180 & 0.0224 & \cellcolor{blue!25}0.2269 & 0.3016 \\
        Tversky &  \cellcolor{blue!25}0.0247 & 0.0285 & \cellcolor{blue!25}0.2131 & 0.2917 \\
        \bottomrule
    \end{tabular}
    \label{tab:brier_scores}
\end{table}

\section{Discussion}
\label{Discuss}


Infiltration of brain tumours beyond their visible boundaries is challenging due to the complex anatomical structure of the brain. Tumour cells often spread along white matter tracts, and certain tumours are likely to affect eloquent connectomes, while others do not. A greater extent of tumour resection is generally associated with an improved prognosis, although this relationship is not strictly linear \cite{weller2021eanom}. The ambiguous region of tumour infiltration presents a significant challenge for neurosurgeons. For example, high-grade gliomas typically exhibit a strong contrast enhancement, often appearing as a hyperintense enhancement ring surrounding a centrally hypointense necrotic core \cite{wen2017response, weller2021eanom}. However, adjacent white and grey matter often display non-enhancing signal alterations. Similarly, low-grade gliomas pose diagnostic difficulties, as their exophytic growth may have indistinct margins with a gradual transition to normal brain tissue \cite{wen2017response}. These signal changes represent a complex interplay of vasogenic edema, microenvironmental alterations, inflammatory responses, and tumour cell infiltration \cite{martin2021advanced}. The distinction between reactive perilesional edema and actual tumour infiltration remains an area of ongoing research and debate \cite{lasocki2019non, martin2021advanced}. Comparable challenges arise in brain metastases \cite{berghoff2013invasion}, dural invasion by low-grade meningiomas \cite{nagashima2006dural}, and aggressive meningiomas infiltrating the brain parenchyma \cite{beutler2024intracranial}.

A clinical imaging perspective of Eq.~\ref{eq:fuzzybound} is provided referencing standard practices in radiology. Fully informed diagnostic decision making with MRI requires multiple sequencing. As stated in \cite{Price2011ImagingBO}, reliance on individual sequences such as T$_1$-weighted or T$_2$-weighted, will underestimate the spread of cancer in the surrounding areas of the determined region of interest. For example, the work in \cite{Stadlbauer2010MagneticRS} measured that the use of proton magnetic resonance spectroscopic imaging, on average, identified tumour margin extensions greater than $20\%$ from what was identifiable using T$_2$ weighted images. We extrapolate this with the case of standard B-mode US. An example of this can be seen in the use of superb microvascular imaging (SMI). In \cite{Cai2024ClinicalAO}, it was seen that the use of SMI, particularly for high-grade gliomas, improved the visualisation of the tumour margin and was associated with an increased likelihood of gross total resection. This clinical perspective further emphasises the limitation of B-mode segmentation and the issue of discretely determining the edge point. The evidence of margin difference when using multiple modalities or sequences indicates not only that discrete margin delineation is not achievable purely using B-mode imaging. However, even with multiple sources of information, the margin uncertainty is only reduced, not eliminated, because of the infiltrative properties.

\section{Conclusion}

The challenge of discretely annotating diffuse, infiltrative, brain tumours in B-mode US is demonstrated. A method for confidence-based annotation is proposed with a protocol derived from computer vision and radiology theory. The method enables sparse and fast annotation which should facilitate rapid dataset generation. The confidence based protocol reduces and/or removes the need for averaging interobserver annotations - through minimising the annotator level epistemic uncertainty by the design of the annotation protocol and method. Evaluation is performed to measure how well the protocol and KDE process can, implicitly, estimate the interobserver variance of 4 professionals, discretely annotating the corresponding images. A reliability diagram is created and an average bin accuracy of 0.1 was measured which indicates correlation between the confidence method and the interobserver variance. In particular, a linear relationship was measured along the tumour margin, with a Pearson correlation of 0.8 and Spearman correlation of 0.75. The downstream application of the confidence annotations is investigated. A UNet is trained for segmentation on both the neuroradiologist discrete annotations (hard label) and the confidence annotations (soft label). Across all evaluation folds, the soft labels produced a better calibrated network with better Brier scores.

The limitations of this study include the potential bias in the analysis of the relationship between confidence annotation and interobserver variation. Although neurosurgeon annotations were not seen during the confidence annotation process, the same authors participated in the prior study \cite{Weld2023ChallengesWS}, which may introduce bias due to the influence of memory. However, this bias was controlled through strict adherence to the annotation protocol and by anchoring the localisation process on top of the neuroradiologist's annotations. Future work to remove interposed normal tissue from the confidence map could be explored, perhaps using sink points. 

\section*{Disclosures}
The authors have no conflict of interest to declare.

\section*{Code, Data, and Materials Availability}
The authors do not have permission to share the data set used in this study. The code can be made available upon reasonable request to the corresponding author.

\section*{Acknowledgments}
This research was supported by the UK Research and Innovation (UKRI) Centre for Doctoral Training in AI for Healthcare (EP/S023283/1), the Royal Society (URF$\setminus $R$\setminus $2 01014]), the NIHR Imperial Biomedical Research Centre, Canon Medical Systems and the Harmsworth Charitable Trust.


\bibliography{report}   
\bibliographystyle{spiejour}   

\end{spacing}
\end{document}